\documentclass[runningheads]{llncs}
\usepackage[T1]{fontenc}
\usepackage{graphicx}
\usepackage{nicematrix}
\begin{document}
\title{Ensemble Learning with Sparse Hypercolumns}
%
%
\author{Julia Dietlmeier\inst{1}\orcidID{0000-0001-9980-0910} \and
Vayangi Ganepola\inst{2}\orcidID{0009-0005-1619-1686} \and
Oluwabukola G. Adegboro\inst{2}\orcidID{0000-0002-2704-8528}
\and \newline
Mayug Maniparambil\inst{1}\orcidID{0000-0002-9976-1920}
\and \newline
Claudia Mazo\inst{3}\orcidID{0000-0003-1703-8964}
\and \newline
Noel E. O'Connor\inst{1}\orcidID{0000-0002-4033-9135}}
\authorrunning{J. Dietlmeier et al.}
%
\institute{$^\text{1}$Insight Research Ireland Centre for Data Analytics, Dublin City University, Glasnevin 9, Dublin, Ireland \\
\email{\{julia.dietlmeier, mayug.maniparambil, noel.oconnor\}@insight-centre.org}
$^\text{2}$Research Ireland Centre for Research Training in Machine Learning (ML-Labs), Dublin City University, Glasnevin 9, Dublin, Ireland\\
\email{\{vayangi.ganepola2, oluwabukola.adegboro2\}@mail.dcu.ie}\\
$^\text{3}$School of Computing, Dublin City University, Glasnevin 9, Dublin, Ireland\\
\email{claudia.mazo@dcu.ie}}
\maketitle              
\begin{abstract}
Directly inspired by findings in biological vision, high-di-mensional hypercolumns are feature vectors built by concatenating multi-scale activations of convolutional neural networks for a single image pixel location. Together with powerful classifiers, they can be used for image segmentation i.e. pixel classification. However, in practice, there are only very few works dedicated to the use of hypercolumns. One reason is the computational complexity of processing concatenated dense hypercolumns that grows linearly with the size $N$ of the training set. In this work, we address this challenge by applying stratified subsampling to the VGG16 based hypercolumns. Furthermore, we investigate the performance of ensemble learning on sparse hypercolumns. Our experiments on a brain tumor dataset show that stacking and voting ensembles deliver competitive performance, but in the extreme low-shot case of $N \leq 20$, a simple Logistic Regression classifier is the most effective method. For 10\% stratified subsampling rate, our best average Dice score is 0.66 for $N=20$. This is a statistically significant improvement of 24.53\% over the standard multi-scale UNet baseline ($p$-value = $[3.07e-11]$, Wilcoxon signed-rank test), which is less effective due to overfitting.

\end{abstract}
\section{Introduction}
The human brain processes vision in layers. It starts with simple details like lines and edges in early brain regions and then builds up to understanding complex objects and scenes in more advanced areas \cite{brain_1982}. Similarly, \textit{hypercolumns} are a computer vision technique that models hierarchical processing of the human brain by integrating information from different layers of a neural network. Image segmentation is the process of assigning a category label to every pixel in an image, effectively treating it as a pixel-by-pixel classification task. Before the advent of the multi-scale skip connection based UNet architecture \cite{UNet_2015}, many image segmentation approaches with Convolutional Neural Networks (CNNs) extracted feature representations from the last layer. This layer, however, may not be optimal because of the coarse spatial resolution. To address this challenge, in 2015 Hariharan et al. \cite{hariharan_cvpr2015} introduced  a hypercolumn descriptor that can be used to classify pixels. It is a feature vector that concatenates the activations of all CNN units directly above a pixel across all layers. This multi-scale approach captures both high-resolution spatial information from early layers and rich semantic information from deeper layers. A typical hypercolumn-based image segmentation pipeline (training phase) consists of creating and concatenating hypercolumns for a number of training images, subsampling, selecting the most relevant features, and finally training a classifier. Specifically, for the case of binary image segmentation with hypercolumns, such classifiers as MLP \cite{skinlesion_hypercolumn_2017,braintumors_miccai2017} and XGB \cite{mitochondria_PRL2019} are reported in the literature.

Generally, combining several classifiers in an ensemble can significantly improve final classification performance in a variety of diverse architectures and applications \cite{ensemble_methods_2012}. The best-known ensemble approaches in machine learning are bagging, boosting, stacking and voting \cite{ensemble_methods_2025}. However, to the best of our knowledge, there are no peer-reviewed and published case studies on ensemble approaches in the context of hypercolumns. There are also very few studies on applying hypercolumns to real-world image segmentation problems \cite{skinlesion_hypercolumn_2017,braintumors_miccai2017,mitochondria_PRL2019}. Our key contributions in this work are as follows:

\begin{itemize}
    \item We develop a hybrid deep and machine learning binary image segmentation pipeline with VGG16 \cite{vgg16_2015} based hypercolumns and ensemble learning
    \item This is the first systematic study investigating ensemble methods (stacking versus voting) to classify sparse multi-scale hypercolumn descriptors in the context of binary image segmentation
    \item This is the first case study quantifying brain tumor segmentation performance by using different stratified \cite{stratify_sampling_2018} hypercolumn subsampling rates
\end{itemize}

\section{Related Work}
There are many multi-scale CNNs, primarily in the domain of semantic segmentation, where they are used to capture both global context and fine-grained spatial detail. These networks generally follow an encoder-decoder architecture with skip connections to share information between different scales. UNet \cite{UNet_2015} and its variants such as, for example, UNet++ \cite{UNet_PP_2018}, and Attention-UNet \cite{Attention_UNet_2018} are some of the most well-known. Transformer-based TransUNet \cite{TranUNet_2024} and SegFormer \cite{Segformer_2021} also apply the multi-scale concept by combining CNNs with attention-based Transformer architectures. While these approaches deliver state-of-the-art segmentation results across a diverse range of datasets, their performance is poor on extremely small datasets. 

Another approach to fine-grained localization is to combine information from multiple scales by using hypercolumns \cite{hariharan_cvpr2015,hariharan_TPAMI2017}. This approach frames image segmentation as a pixel classification problem. It uses a fixed pretrained feature extractor and a simple classifier, bypassing the need to train a large deep neural network from scratch. Therefore it is much more robust against overfitting on limited data \cite{hariharan_TPAMI2017,mitochondria_PRL2019}.

The hypercolumns-based approach has found its applications in biomedical and medical imaging, such as for the segmentation of mitochondria \cite{mitochondria_PRL2019}, skin lesions \cite{skinlesion_hypercolumn_2017}, and brain tumors \cite{braintumors_miccai2017}. Despite these few works, its use remains largely underexplored. One of the reasons is the computational cost of processing $N$ concatenated hypercolumns for a training set consisting of $N$ images. In this work we address this challenge and propose a processing pipeline that not only creates sparse hypercolumns but also leverages the strengths of ensemble learning.

\section{Architecture}
\begin{figure*}[!t]
\includegraphics[width=\textwidth]{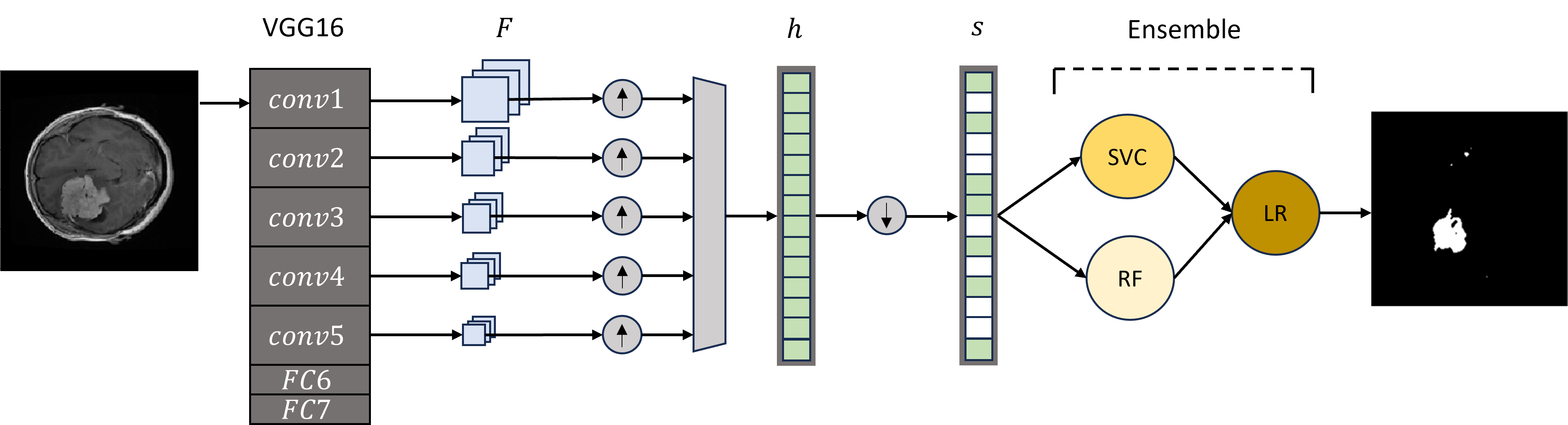}
\caption{Our hypercolumn-based processing pipeline for  binary pixel classification. } \label{fig1}
\end{figure*}

Our processing pipeline is illustrated in Figure \ref{fig1}.
In order to construct the hypercolumns, we first extract features $F$ from all five convolutional blocks of a pretrained (on the ImageNet dataset) VGG16 network \cite{vgg16_2015}. The size of feature maps decreases with the VGG16 depth because of the max pooling layers. Therefore, we apply bilinear upsampling to upsample multi-scale feature maps to the input resolution of $224 \times 224$ pixels. We concatenate the corresponding feature vectors along the channel dimension and flatten to obtain a dense hypercolumn $h$. This results in a hypercolumn with the dimension $50176 \times 4964$ for one $224 \times 224$ image. 

In order to construct a training set consisting of $N$ images, we need to concatenate the $N$ hypercolumns along the row dimension, forming a ($N \times 50176) \times 4964$ matrix. The computational complexity increases with $N$ and therefore we need to subsample the data. Specifically, we apply stratified subsampling to obtain a sparse hypercolumn $s$. The motivation to use stratified subsampling instead of simple random subsampling is that the foreground labels (white pixels in Figure \ref{fig1}) comprise a minority class. In this case, stratification ensures accurate representation of pixel-level labels within a sample. This approach prevents the model from ignoring the rare but important foreground pixels, which can happen with random subsampling \cite{stratify_sampling_2018}. By balancing the representation of both foreground and background classes, stratified sampling helps to train a more robust and effective segmentation model.

For a number $N$ of training images we combine (concatenate) the $N$ dense hypercolumns first and then perform subsampling, ensuring the sample reflects the entire dataset's distribution.

Sparse combined hypercolumns are then fed into an ensemble of base classifiers. We compare stacking versus voting ensembles. The main difference is in how these ensembles combine the predictions of their base models. Stacking uses a meta-learner and voting uses simple aggregation methods such as averaging or majority voting \cite{ensemble_methods_2012,ensemble_methods_2025}. For the stacking ensemble we are using an ensemble of RandomForestClassifier (RF), Linear SupportVectorClassifier (LinearSVC), and LogisticRegression (LR) as a meta-learner. The soft voting ensemble we use is a combination of RF, non-linear SVC and LR base classifiers. 

\section{Experiments}

We evaluate the proposed image segmentation pipeline on the Cheng et al. dataset \cite{Cheng_dataset} of brain tumors\footnote{available on https://figshare.com/articles/dataset/brain\_tumor\_dataset/1512427}. This open-source dataset consists of 3064 2D T1-weighted contrast-enhanced MRI scans with three brain tumor types such as meningioma, glioma and pituitary. Specifically, in this work, we focus on the pathology of the meningioma tumor class. 

We use the train and test split from \cite{Grace_IMVIP2024} and obtain a meningioma training set of 496 images and a test set of 140 images. The input resolution of images varies and therefore we resize all images to $224 \times 224$ as input to the VGG16-based hypercolumn model. We normalize the images for the hypercolumn model with the statistics of the ImageNet dataset. We do not apply any data augmentation to the training set.

In order to construct a dense hypercolumn for $N \ll 496$ we first randomly sample the training set and concatenate the $N$ hypercolumns obtained. Then we apply stratified subsampling to obtain the sparse combined hypercolumn. To ensure the robustness of our findings, we conduct five experimental runs. Each run utilizes a different training subset $N$, which is obtained through random sampling of the full training set. The final results are presented as the average of these runs, an approach consistent with the Monte Carlo strategy \cite{monte_carlo_1995}. We investigate six models in total  - stacking, voting, LR, RF, SVC, and UNet. All models are evaluated on exactly the same randomly sampled training subsets.

\subsection{Implementation and Training Details}
The processing pipeline was implemented in Python 3.9, PyTorch 2.0.1, and using the \textit{scikit-learn} library. We set random seed of 42 for reproducibility. All experiments were performed on a desktop computer with the Ubuntu operating system 18.04.3 LTS with the Intel(R) Core(TM) i9-9900K CPU, Nvidia GeForce RTX 2080 Ti GPU, and a total of 62GB RAM.

The parameter settings for the classifiers used are shown in Table~\ref{tab1}. We used \textit{soft} voting with weights $[0.4, 0.4, 0.2]$ for the RF, SVC and LR respectively. For the stacking ensemble we used the LinearSVC because it is much faster and more scalable than its non-linear SVC version.

\begin{table}[h]
\caption{Base classifier configuration.}\label{tab1}
\centering
\begin{tabular}{|l|l|l|}
\hline
Base classifiers &  Number of estimators & Max iterations\\
\hline
Random Forest Classifier (RF) &  10 & n/a\\
Linear Support Vector Classifier (SVC) &  n/a & 1000\\
Logistic Regression Classifier (LR) & n/a & 1000\\
\hline
\end{tabular}
\end{table}
The vanilla UNet model consists of three encoder (downsampling path) and three decoder (upsampling path) convolutional blocks, bottleneck convolutional block, and skip connections. Batch Normalization layers are added after each Conv2d layer. The number of feature maps increases from 64 to 512 with the depth of the network. The last Conv2d layer has 1 output channel because of the binary segmentation case and the kernel size of 1. We train the UNet baseline from scratch for 100 epochs using Binary Cross Entropy with Logits loss, the Adam optimizer with the learning rate of $1e-4$ and the batch size of $2$.

\subsection{Quantitative Results}
\begin{figure*}[!t]
\includegraphics[width=\textwidth]{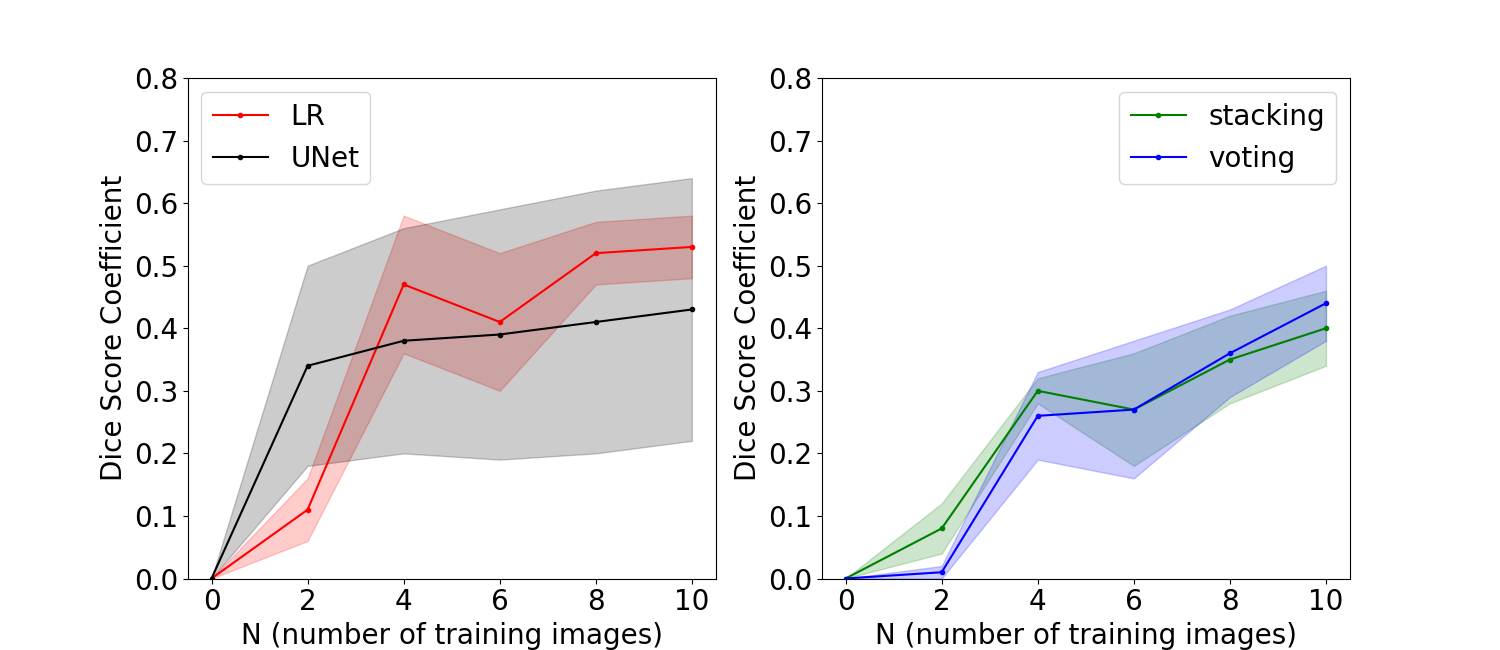}
\caption{Dice score coefficient performance of LR versus UNet (left) and stacking versus voting (right). All models were trained on the same amount of images for 1\% stratified subsampling rate. Graphs are showing mean (lines) $\pm$ standard deviation (filled areas). } \label{fig2}
\end{figure*}

\begin{figure*}[!t]
\centering
\includegraphics[width=\textwidth]{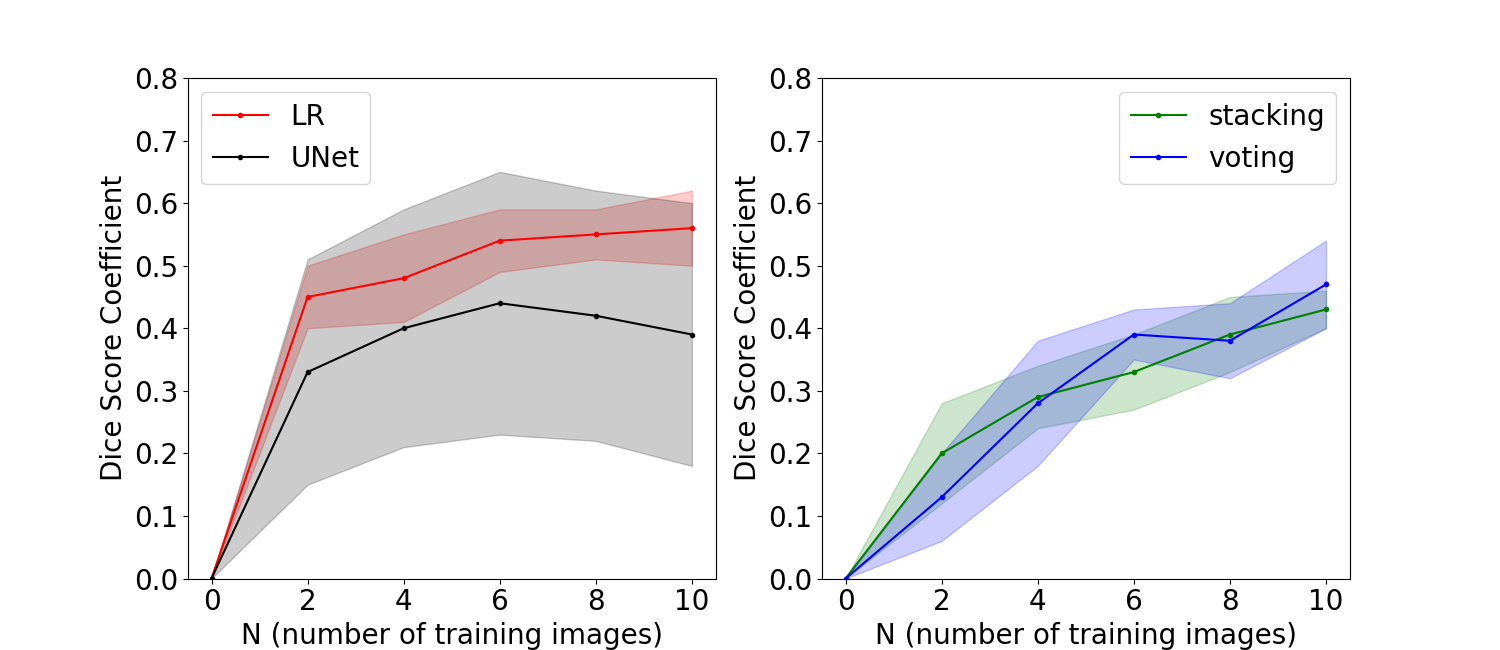}
\caption{Dice score coefficient performance of LR versus UNet (left) and stacking versus voting (right). All models were trained on the same amount of images for 10\% stratified subsampling rate. Graphs are showing mean (lines) $\pm$ standard deviation (filled areas). } \label{fig3}
\end{figure*}

Due to the computational and time related constraints we do not perform the hypercolumn based experiments on the full training set but rather in the low-data regime for $N\leq 20$. For the evaluation we apply standard pixel-level segmentation metrics such as accuracy, precision, recall, Jaccard and Dice coefficients. Selected results are provided in Tables \ref{tab2} and  \ref{tab3} and Figures \ref{fig2} and \ref{fig3}. Statistical test results (Wilcoxon signed-rank test) are shown below in Table~\ref{tab_pvalues}.
\begin{table}[h]
\caption{Comparison between the hypercolumn+LR and the UNet models. Dice performance gains are computed based on the mean values. }\label{tab_pvalues}
\centering

\begin{NiceTabular}{cccccc}[hvlines]
Subsampling rate & N & hypercolumn+LR $\uparrow$ & UNet $\uparrow$ & Dice gain $\uparrow$  & $p$-value $\downarrow$
 \\
\Block{3-1}{1\% } & 2 & 0.11 $\pm$ 0.05 & 0.34 $\pm$ 0.16 & $-67.65$ \% &  $2.23e-05$  \\

                               & 10 & 0.53 $\pm$ 0.05 & 0.43 $\pm$ 0.21 & $23.26$ \% &  $1.59e-07$  \\
                               
                               & 20 & 0.61 $\pm$ 0.03 & 0.56 $\pm$ 0.26 & $8.93$ \% & 0.663   \\
\hline
\Block{3-1}{10\% } & 2 & 0.45 $\pm$ 0.05 & 0.33 $\pm$ 0.18 & $36.36$ \% & $1.52e-14$   \\
                               & 10 & 0.56 $\pm$ 0.06 & 0.39 $\pm$ 0.21 & $43.59$ \% & $1.40e-20$   \\
                               & 20 & 0.66 $\pm$ 0.02 
 & 0.53 $\pm$ 0.25  & 24.53 \% &  $3.07e-11$  \\
\hline

\end{NiceTabular}
\end{table}


\begin{table*}[!t]
\caption{Selected quantitative results (mean and standard deviation) for the case of 1\% stratified subsampling rate. Best results are in red and second best in blue.}\label{tab2}
\centering

\begin{NiceTabular}{lcccccc}[hvlines]
 & Models & Accuracy$\uparrow$ & Precision$\uparrow$ & Recall$\uparrow$ & Jaccard$\uparrow$ & Dice$\uparrow$ \\
\Block{6-1}{\rotate N=2} & hypercolumn+LR& 0.98 $\pm$ 0.00 & 0.35 $\pm$ 0.12 & 0.08 $\pm$ 0.04 & 0.08 $\pm$ 0.04 & \textcolor{blue}{0.11 $\pm$ 0.05}  \\

                               & hypercolumn+RF & 0.98 $\pm$ 0.00 & 0.04 $\pm$ 0.03 & 0.00 $\pm$ 0.00 & 0.00 $\pm$ 0.00 & 0.00 $\pm$ 0.00  \\
                               
                               & hypercolumn+SVC & 0.98 $\pm$ 0.00 & 0.02 $\pm$ 0.01 & 0.00 $\pm$ 0.00 & 0.00 $\pm$ 0.00 & 0.00 $\pm$ 0.00  \\
                               
                               & hypercolumn+stacking & 0.93 $\pm$ 0.04 & 0.08 $\pm$ 0.05 & 0.20 $\pm$ 0.12 & 0.05 $\pm$ 0.03 & 0.08 $\pm$ 0.04  \\
                               
                               & hypercolumn+voting & 0.98 $\pm$ 0.00 & 0.08 $\pm$ 0.04 & 0.01 $\pm$ 0.01 & 0.01 $\pm$ 0.01 & 0.01 $\pm$ 0.01  \\
                               
                               & UNet & 0.79 $\pm$ 0.38 & 0.40 $\pm$ 0.20 & 0.57 $\pm$ 0.23 & 0.25 $\pm$ 0.12 & \textcolor{red}{0.34 $\pm$ 0.16}  \\
\hline
\Block{6-1}{\rotate N=10} & hypercolumn+LR& 0.99 $\pm$ 0.00 & 0.75 $\pm$ 0.03 & 0.49 $\pm$ 0.08 & 0.42 $\pm$ 0.04 & \textcolor{red}{0.53 $\pm$ 0.05}  \\
                               & hypercolumn+RF & 0.98 $\pm$ 0.00 & 0.41 $\pm$ 0.05 & 0.08 $\pm$ 0.05 & 0.08 $\pm$ 0.04 & 0.11 $\pm$ 0.06  \\
                               & hypercolumn+SVC & 0.99 $\pm$ 0.00 & 0.60 $\pm$ 0.07 & 0.22 $\pm$ 0.07 & 0.21 $\pm$ 0.07 & 0.29 $\pm$ 0.08  \\
                               & hypercolumn+stacking & 0.98 $\pm$ 0.01 & 0.51 $\pm$ 0.05 & 0.43 $\pm$ 0.10 & 0.29 $\pm$ 0.05 & 0.40 $\pm$ 0.06  \\
                               & hypercolumn+voting & 0.99 $\pm$ 0.01 & 0.75 $\pm$ 0.05 & 0.37 $\pm$ 0.06 & 0.35 $\pm$ 0.05 & \textcolor{blue}{0.44 $\pm$ 0.06}  \\
                               & UNet & 0.80 $\pm$ 0.38 & 0.52 $\pm$ 0.25 & 0.62 $\pm$ 0.21 & 0.36 $\pm$ 0.18 & 0.43 $\pm$ 0.21  \\
\hline
\Block{6-1}{\rotate N=20} & hypercolumn+LR& 0.99 $\pm$ 0.00 & 0.77 $\pm$ 0.02 & 0.59 $\pm$ 0.05 & 0.49 $\pm$ 0.02 & \textcolor{red}{0.61 $\pm$ 0.03}  \\
                               & hypercolumn+RF & 0.99 $\pm$ 0.00 & 0.48 $\pm$ 0.08 & 0.10 $\pm$ 0.05 & 0.10 $\pm$ 0.04 & 0.15 $\pm$ 0.06  \\
                               & hypercolumn+SVC & 0.99 $\pm$ 0.00 & 0.77 $\pm$ 0.03 & 0.35 $\pm$ 0.08 & 0.33 $\pm$ 0.06 & 0.43 $\pm$ 0.07  \\
                               & hypercolumn+stacking & 0.99 $\pm$ 0.00 & 0.63 $\pm$ 0.04 & 0.54 $\pm$ 0.06 & 0.39 $\pm$ 0.02 & 0.51 $\pm$ 0.02  \\
                               & hypercolumn+voting & 0.99 $\pm$ 0.00 & 0.84 $\pm$ 0.02 & 0.49 $\pm$ 0.07 & 0.45 $\pm$ 0.04 & \textcolor{blue}{0.56 $\pm$ 0.05}  \\
                               & UNet & 0.81 $\pm$ 0.38 & 0.65 $\pm$ 0.32 & 0.72 $\pm$ 0.14 & 0.48 $\pm$ 0.23 & {0.56 $\pm$ 0.26}  \\

\hline
\end{NiceTabular}
\end{table*}

\begin{table*}[!t]
\caption{Selected quantitative results (mean and standard deviation) for the case of 10\% stratified subsampling rate. Best results are in red and second best in blue.}\label{tab3}
\centering

\begin{NiceTabular}{lcccccc}[hvlines]
 & Models & Accuracy$\uparrow$ & Precision$\uparrow$ & Recall$\uparrow$ & Jaccard$\uparrow$ & Dice$\uparrow$ \\
\Block{6-1}{\rotate N=2} & hypercolumn+LR& 0.99 $\pm$ 0.00 & 0.69 $\pm$ 0.03 & 0.39 $\pm$ 0.06 & 0.34 $\pm$ 0.05 & \textcolor{red}{0.45 $\pm$ 0.05}  \\
                               & hypercolumn+RF & 0.98 $\pm$ 0.00 & 0.21 $\pm$ 0.12 & 0.03 $\pm$ 0.03 & 0.03 $\pm$ 0.05 & 0.05 $\pm$ 0.05  \\
                               & hypercolumn+SVC & 0.98 $\pm$ 0.00 & 0.27 $\pm$ 0.13 & 0.06 $\pm$ 0.03 & 0.06 $\pm$ 0.03 & 0.08 $\pm$ 0.04  \\
                               & hypercolumn+stacking & 0.94 $\pm$ 0.02 & 0.18 $\pm$ 0.06 & 0.41 $\pm$ 0.12 & 0.13 $\pm$ 0.05 & 0.20 $\pm$ 0.08  \\
                               & hypercolumn+voting & 0.99 $\pm$ 0.00 & 0.36 $\pm$ 0.15 & 0.10 $\pm$ 0.06 & 0.10 $\pm$ 0.06 & 0.13 $\pm$ 0.07  \\
                               & UNet & 0.79 $\pm$ 0.39 & 0.41 $\pm$ 0.21 & 0.57 $\pm$ 0.29 & 0.26 $\pm$ 0.14 & \textcolor{blue}{0.33 $\pm$ 0.18}  \\
\hline
\Block{6-1}{\rotate N=10} & hypercolumn+LR& 0.99 $\pm$ 0.00 & 0.73 $\pm$ 0.03 & 0.53 $\pm$ 0.08 & 0.44 $\pm$ 0.04 & \textcolor{red}{0.56 $\pm$ 0.06}  \\
                               & hypercolumn+RF & 0.98 $\pm$ 0.00 & 0.45 $\pm$ 0.10 & 0.07 $\pm$ 0.05 & 0.07 $\pm$ 0.04 & 0.11 $\pm$ 0.06  \\
                               & hypercolumn+SVC & 0.99 $\pm$ 0.00 & 0.78 $\pm$ 0.07 & 0.37 $\pm$ 0.06 & 0.36 $\pm$ 0.05 & 0.45 $\pm$ 0.06  \\
                               & hypercolumn+stacking & 0.98 $\pm$ 0.00 & 0.46 $\pm$ 0.04 & 0.55 $\pm$ 0.09 & 0.32 $\pm$ 0.02 & 0.43 $\pm$ 0.03  \\
                               & hypercolumn+voting & 0.99 $\pm$ 0.00 & 0.79 $\pm$ 0.08 & 0.40 $\pm$ 0.07 & 0.38 $\pm$ 0.06 & \textcolor{blue}{0.47 $\pm$ 0.07}  \\
                               & UNet & 0.90 $\pm$ 0.17 & 0.54 $\pm$ 0.27 & 0.36 $\pm$ 0.21 & 0.32 $\pm$ 0.18 & 0.39 $\pm$ 0.21  \\
\hline
\Block{6-1}{\rotate N=20} & hypercolumn+LR& 0.99 $\pm$ 0.00 & 0.79 $\pm$ 0.04 & 0.65 $\pm$ 0.04 & 0.54 $\pm$ 0.02 & \textcolor{red}{0.66 $\pm$ 0.02}  \\
                               & hypercolumn+RF & 0.99 $\pm$ 0.00 & 0.62 $\pm$ 0.04 & 0.12 $\pm$ 0.03 & 0.12 $\pm$ 0.03 & 0.18 $\pm$ 0.04  \\
                               & hypercolumn+SVC & 0.99 $\pm$ 0.00 & 0.88 $\pm$ 0.02 & 0.52 $\pm$ 0.03 & 0.50 $\pm$ 0.02 & 0.61 $\pm$ 0.02  \\
                               & hypercolumn+stacking & 0.98 $\pm$ 0.00 & 0.58 $\pm$ 0.04 & 0.55 $\pm$ 0.05 & 0.39 $\pm$ 0.03 & 0.50 $\pm$ 0.04  \\
                               & hypercolumn+voting & 0.99 $\pm$ 0.00 & 0.89 $\pm$ 0.02 & 0.53 $\pm$ 0.02 & 0.50 $\pm$ 0.01 & \textcolor{blue}{0.61 $\pm$ 0.01}  \\
                               & UNet & 0.81 $\pm$ 0.36 & 0.62 $\pm$ 0.31 & 0.71 $\pm$ 0.15 & 0.46 $\pm$ 0.22 & 0.53 $\pm$ 0.25  \\

\hline
\end{NiceTabular}
\end{table*}

It can be seen from Table~\ref{tab_pvalues} that the hypercolumn-LR model outperforms all other models including the baseline UNet (trained on the same amount of training images) in all cases with an exception for $N=2$ (1\% subsampling rate). The $p$-value for $N=20$ indicates that the difference in performance is not statistically significant. However, the results are remarkable given the fact that the sparse hypercolumn based approach works with only $1\%$ to $10\%$ of the data while the UNet based approach utilizes the full $100\%$ image information. We also highlight that the Dice values obtained are generally lower as compared to training on the full meningioma set of $N=496$. For example, the vanilla UNet model achieves Dice=0.89 on the full training set. 

The subsampling rate of 10\% delivers better results than that of 1\%, which makes sense as we have more data points available to build hypercolumns with the decreasing sparsity. It can be observed that UNet based results have higher standard deviation than the hypercolumn-based approach with base and ensemble classifiers. This indicates that UNet's results are not as accurate and repeatable. We think that this is because of the small training set size where the effect of the Monte Carlo sampling is amplified. That means that the high variance in the results is a direct consequence of the sampling uncertainty being amplified by the small size of the dataset \cite{stats_learning_book}.

We also perform statistical (Wilcoxon signed-rank test) tests on the results obtained. For example, our best Dice=0.66 is obtained with LR-hypercolumn model for $N=20$ and 10\% subsampling rate. UNet achieves only Dice=0.53 with the same configuration. This corresponds to the statistically significant performance gain of 24.53\% with a $p$-value$=3.07e-11$.

Figures \ref{fig2} and \ref{fig3} reveal that stacking and voting deliver competitive performance for both 1\% and 10\% subsampling rates.

\subsection{Qualitative Results}
\begin{figure*}[!t]
\includegraphics[width=\textwidth]{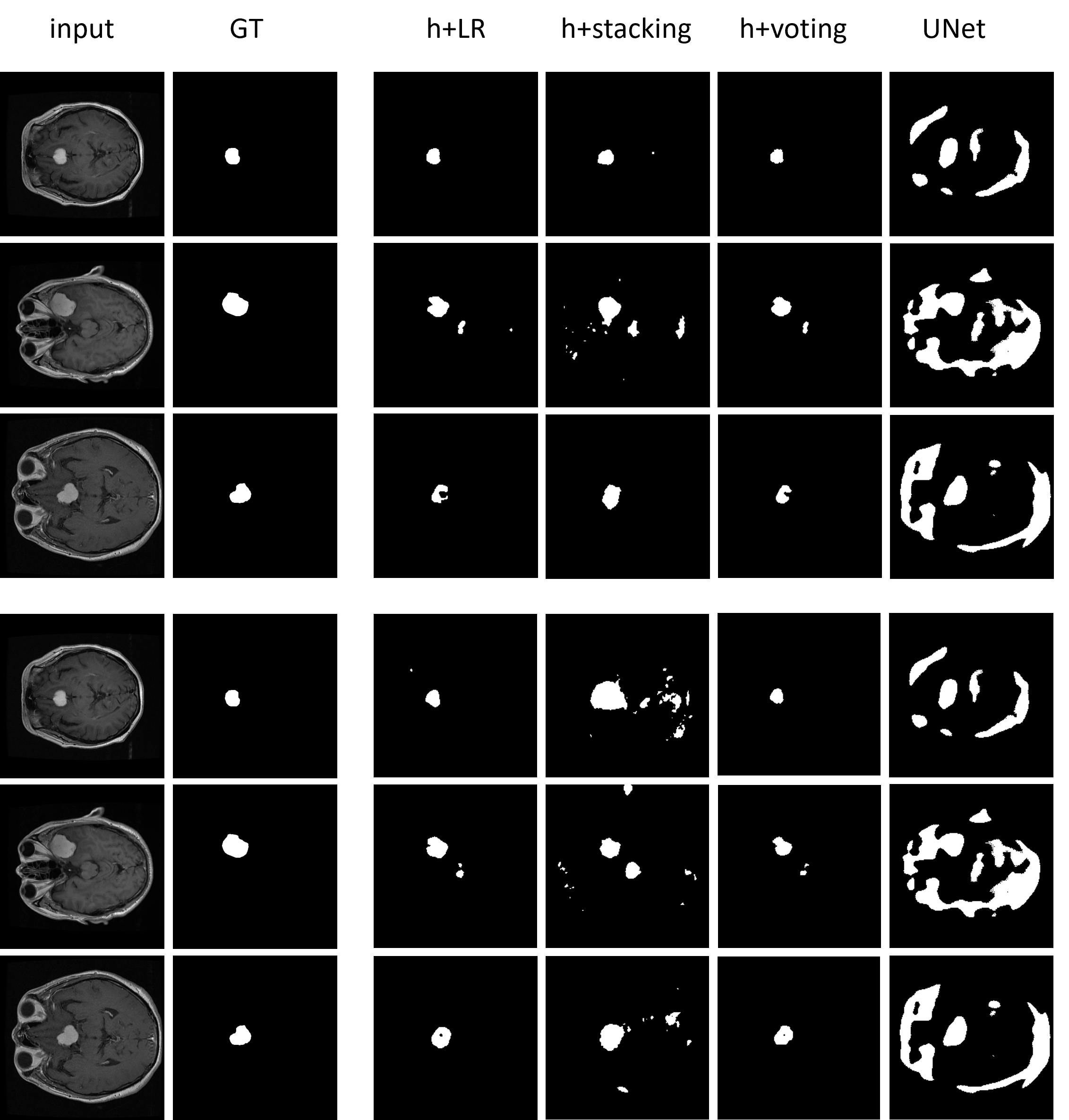}
\caption{Selected qualitative results for $N=10$ training images. The first three rows show results for 1\% subsampling rate and the last three rows show the results for 10\% subsampling rate. Note that UNet is trained on $N=10$ images without subsampling. } \label{fig_qualitative}
\end{figure*}

Qualitative results are shown in Figure~\ref{fig_qualitative} and provide the comparison between the ground truth (GT), hypercolumn+LR (h+LR), hypercolumn+stacking (h+ stacking), hypercolumn+voting (h+voting) and UNet models. The most striking observation is the number of false positives in the predictions returned by the UNet model where it incorrectly identifies parts of the background as part of the tumor. These are areas that were segmented by UNet but are not present in the ground truth. For both, the 1\% and 10\% subsampling cases, all three hypercolumn-based models shown qualitatively outperform the UNet model. 

Interestingly, the stacking ensemble qualitatively underperfoms in 10\% subsampling case (in these selected samples) compared to the 1\% subsampling case. 

\subsection{Computational Details}
Computational details are provided in Table~\ref{tab4}.
\begin{table}[h]
\caption{Computational details. The total number of parameters for base and ensemble classifiers are obtained after training on $N=20$ images with the 10\% subsampling rate. Inference time is given per image. Times are in seconds. }\label{tab4}
\centering
\begin{tabular}{|l|c|c|c|c|}
\hline
Models &  Total parameters & Inference time & Training time & Device\\
\hline
hypercolumn model & 14,714,688 & - & - & GPU \\
+LR & 1,473 & 0.48 & 7.66 & CPU\\
+RF &  5,290 & 0.17 & 13.56 & CPU\\
+SVC &  2,536,257 & 63.38 & 1062.97 & CPU\\
+stacking & 1,486 & 0.61 & 254.95 & CPU\\
+voting & 2,886,604 & 80.40 & 1347.26 & CPU\\
UNet & 7,701,825 & 0.001 & 38.86 & GPU\\

\hline
\end{tabular}
\end{table}
We observe that the voting ensemble classifier which contains the non-linear SVC is slower in inference time as compared to the stacking ensemble which contains its linear SVC version. The reason for this is the optimization process of SVC that has a computational complexity of $O(N^2)$ or $O(N^3)$ and its prediction time directly depends on the number of support vectors it has learned during training \cite{SVC_slow}. In contrast, linear SVC needs only to compute a single dot product and therefore has a computational complexity of $O(N)$. This fact makes this classifier more suitable for processing large hypercolumns.  
\section{Conclusion}
Hypercolumns offer a powerful computational parallel to how the human brain integrates information from different levels of its visual hierarchy to achieve the pixel-level understanding of a scene. Our case study on binary image segmentation with hypercolumns in the extreme low-shot data scenario revealed that a simple Logistic Regression classifier outperformed the ensemble methods such as stacking and voting. However, we did not investigate the subsampling rates of $>10\%$ which could lead to higher accuracy of ensembles versus linear classifiers. It is well known that overfitting is a major weakness of deep learning models when there is not enough data to train on. In this context, our results demonstrate that incorporating hypercolumns leads to statistically significant improvements in segmentation performance over the standard UNet architecture. Our future research will be dedicated to investigating other hypercolumn subsampling approaches, such as the information-theoretic ones \cite{conclusion_infotheoretic_2024}, to more effectively utilize the information present in the training images.

\begin{credits}
\subsubsection{\ackname} This work was funded by the Insight Research Ireland Centre for Data Analytics and
partly supported by Taighde \'Eireann – Research Ireland under Grant number 18/CRT/6183
(Ganepola, Adegboro).

\subsubsection{\discintname}
The authors have no competing interests to declare that are
relevant to the content of this article. 
\end{credits}
%
%
%
\bibliographystyle{splncs04}
\bibliography{main}
\end{document}